\documentclass[final,3p,times,twocolumn]{elsarticle}

\usepackage{url}

\usepackage{amssymb}
\usepackage[colorlinks=true]{hyperref}
\usepackage{graphicx}
\usepackage{url}
\usepackage{subcaption}
\usepackage{lscape}
\usepackage{amsmath}
\usepackage{multirow}
\usepackage{xcolor} 
\usepackage{caption}
\usepackage{adjustbox}
\usepackage[figuresright]{rotating}
\usepackage{algorithm} 
\usepackage{algpseudocode} 
\usepackage{comment}
\usepackage{booktabs} 
\usepackage{rotating} 

\biboptions{numbers,sort&compress}
\setcitestyle{numbers,square,comma,sort&compress}

\usepackage{calc}
\makeatletter
\def\ps@pprintTitle{%
  \let\@oddhead\@empty
  \let\@evenhead\@empty
  \def\@oddfoot{\reset@font\hfil\thepage\hfil}
  \let\@evenfoot\@oddfoot
}
\makeatother

\begin{document}

\begin{frontmatter}
\title{Detection of On-Ground Chestnuts Using Artificial Intelligence Toward Automated Picking}
\author[label1]{Kaixu Fang}
\author[label2]{Yuzhen Lu\corref{cor1}}
\cortext[cor1]{Corresponding author}
\ead{luyuzhen@msu.edu}
\author[label2]{Xinyang Mu}

\address[label1]{School of Automation Engineering, University of Electronic Science and Technology of China, Chengdu, Sichuan 611731, China}
\address[label2]{Department of Biosystems \& Agricultural Engineering, Michigan State University, East Lansing, MI 48824, United States}

\begin{abstract}
Traditional mechanized chestnut harvesting is too costly for small producers, non-selective, and prone to damaging nuts. Accurate, reliable detection of chestnuts on the orchard floor is crucial for developing low-cost, vision-guided automated harvesting technology. However, developing a reliable chestnut detection system faces challenges in complex environments with shading, varying natural light conditions, and interference from weeds, fallen leaves, stones, and other foreign on-ground objects, which have remained unaddressed. This study collected 319 images of chestnuts on the orchard floor, containing 6524 annotated chestnuts. A comprehensive set of 29 state-of-the-art real-time object detectors, including 14 in the YOLO (v11-13) and 15 in the RT-DETR (v1-v4) families at varied model scales, was systematically evaluated through replicated modeling experiments for chestnut detection. Experimental results show that the YOLOv12m model achieves the best mAP@0.5 of 95.1\% among all the evaluated models, while the RT-DETRv2-R101 was the most accurate variant among RT-DETR models, with mAP@0.5 of 91.1\%. In terms of mAP@[0.5:0.95], the YOLOv11x model achieved the best accuracy of 80.1\%. All models demonstrate significant potential for real-time chestnut detection, and YOLO models outperformed RT-DETR models in terms of both detection accuracy and inference, making them better suited for on-board deployment. Both the dataset and software programs in this study have been made publicly available at \url{https://github.com/AgFood-Sensing-and-Intelligence-Lab/ChestnutDetection}.
\end{abstract}


\end{frontmatter}


\graphicspath{{benchmark_figures/Chestnut_figures/}}

\section{Introduction}

Chestnut (Castanea spp.) is a nutritious and popular nut crop, rich in vitamins and minerals, low in fat, and high in dietary fiber. Its sweet flavor is particularly favored by American consumers. According to the 2022 USDA Census of Agriculture \cite{census2022available}, 2,845 growers in the United States (U.S.) manage a total of 10,049 acres of fruiting and non-fruiting chestnut orchards, with an average farm size of 3.5 acres. Farms with a size of 2 hectares (4.94 acres) or less and fewer than 600 chestnut trees are considered small-scale farms \cite{kang2008development}, indicating a significant small-scale production pattern in chestnut cultivation in the U.S. Michigan is a major chestnut-producing state in the U.S., accounting for approximately 13\% of the national planted area, of which approximately 35\% is non-fruiting chestnut orchards. A large amount of land resources has not yet been fully converted into actual production capacity, and the industry has huge room for development and improvement.

Chestnuts are highly seasonal fruits that can only maintain their peak commercial quality, size, and health for a relatively short period of time \cite{massantini2021evaluating}. One of the primary challenges in chestnut production is its susceptibility to pest damage and quality degradation, which often leads to significant post-harvest loss. During the harvest period, chestnuts naturally fall from protective shells  (known as burrs) to the orchard floor. Once in direct contact with the ground, the nuts are exposed to adverse environmental conditions, including soil, fallen branches and leaves, surface microbial communities, precipitation, fluctuations in temperature and humidity, etc. Fungi have been identified as the main cause of postharvest chestnut decay \cite{jermini2006influence}. Fungal infection typically occurs after nuts fall to the ground and come into contact with soil, plant residues, and/or dirty water \cite{lee2016efficacy}. Additionally, fallen chestnuts are often subjected to vibration or friction, which can result in micro-cracks or abrasions on the shell or pericarp. These micro-wounds serve as the entry points for fungal spores, mycelium, or small insects. Damage caused by wildlife foraging further exacerbates yield and quality losses. Consequently, if the frequency of harvesting and picking is not high enough, chestnuts must be harvested promptly after falling \cite{sieber2007effects} to minimize ground exposure and risk of nut deterioration and loss, and ensure optimal product quality.

Despite these risks, on-ground chestnut harvesting still relies primarily on manual picking, which is highly labor-intensive, time-consuming, and increasingly unsustainable as orchard acreage expands. As the chestnut industry continues to grow—particularly among small and family-operated farms—producers are often faced with harvest volumes that exceed what a limited workforce can reasonably manage. To reduce labor demands, various mechanical harvesting systems have been introduced; however, these solutions only partially address the challenges of efficiency, labor dependence, and nut quality preservation. Currently available mechanical harvesting equipment mainly includes vacuum-based harvesters and mechanical sweepers, which may be trailed, mounted, or self-propelled depending on orchard conditions and operational requirements \cite{sieber2007effects}. Vacuum harvesters,  commonly used on uneven or sloping terrain, collect chestnuts from the orchard floor through suction pipes and deposit them into collection bags, whereas mechanical sweepers gather windfallen nuts using rotating brushes and conveyor belts.

Although these machines can reduce some manual effort, they still require continuous human involvement for driving, monitoring, and post-harvest handling, and their high acquisition costs (e.g., \$50,000 - \$100,000 per unit) limit adoption by small-scale producers. Moreover, compared with manual picking, mechanical harvesting has been shown to increase the risk of physical damage to chestnuts, including bruising, abrasions, kernel darkening, and off-odor development \cite{guyer20122012,colantoni2014carletti}. In vacuum-based systems, for example, internal kernel damage caused by suction forces may not be immediately visible at harvest and often becomes apparent during storage, leading to increased postharvest losses and the need for additional handling or treatments \cite{ekman2021improved}. While such systems are often described as "automated," they do not eliminate labor-intensive steps such as separation, quality inspection, and rehandling, nor do they substantially reduce overall labor costs. These limitations underscore the need not merely for improved mechanization, but for a genuinely autonomous harvesting solution that minimizes labor input while preserving nut quality.

In recent years, several studies have explored small- to medium-scale, cost-effective harvesting assistance systems for chestnut production. Kang \& Guyer (2008) \cite{kang2008development} developed and evaluated three chestnut harvester prototypes and proposed a venturi-based separation device to distinguish chestnuts from empty shells. Among them, the airlock blade system successfully picked up all the scattered material with a picking efficiency of about 56 kg/h. However, the harvesting performance was inconsistent, and the system remained relatively bulky and energy-intensive. De Kleine \& Guyer (2013) \cite{de2013design} introduced an airflow-adjustable harvesting system capable of collecting and separating chestnuts from orchard debris, thereby improving the operational efficiency of small orchards. According to tests, the highest chestnut harvesting efficiency could reach 88.44\%. It is noted that the performance of the harvesting system was strongly affected by the nut-to-debris ratio and material feed rate. More recently, Greg Peck and colleagues at Cornell AgriTech demonstrated the Silverfox Harvester, which can collect up to 800 pounds of chestnuts per hour at a cost of less than \$4,000, offering an affordable option for small producers \cite{producetech2026silverfox}. Nevertheless, this system still relies on manual operation and does not support autonomous harvesting. Overall, existing small-scale mechanical harvesting systems remain constrained by labor dependence, incomplete automation, performance reliability, and the need for additional separation and processing steps.

To address these challenges, there is a clear need to develop a low-cost, high-precision autonomous chestnut harvesting system that incorporates vision capabilities to identify chestnuts on the orchard floor and accurately collect them without causing damage. Such a system would eliminate downstream separation processes, significantly reduce labor requirements, and enable efficient, high-quality harvesting tailored for small-scale chestnut producers.

As the core component of an intelligent harvesting platform, reliable machine vision is fundamental to accurate perception, decision-making, and robotic operation. Advances in machine learning (ML) and artificial intelligence (AI)-based vision technologies have significantly accelerated progress in agricultural automation. Early applications of ML focused on crop monitoring and yield prediction \cite{rashid2021a}, and have since expanded to a wide range of tasks. AI vision systems have greatly improved the efficiency of pest and disease identification, crop health monitoring, and phenotypic analysis, enabling rapid, data-driven management decisions without human intervention \cite{batz2023from}. In harvesting automation, Alaaudeen et al. (2024) \cite{alaaudeen2024intelligent}, for example, combined computer vision with robot harvesting to realize autonomous apple picking, reporting recognition success rates exceeding 95\% and the retry rates below 12\%. Recent studies have also demonstrated the effectiveness of deep learning in chestnut-related vision tasks. Adão et al. (2019) \cite{ado2019hru} successfully classified and segmented chestnuts using convolutional neural networks (CNN), achieving a classification accuracy of 91\%. Sun et al. (2023) \cite{sun20232023} applied semantic segmentation to UAV images for chestnut tree cover detection, obtaining an average F1 score of 86.13\%. However, traditional machine learning and early deep learning approaches often exhibit limited generalization under the complex visual conditions encountered in real chestnut orchard environments. Challenges such as variable illumination, occlusion, orchard floor clutter, and dynamic environmental conditions frequently degrade detection accuracy as well as real-time performance, limiting their practical deployment in autonomous harvesting systems.

The first step toward automated chestnut harvesting is the development of a robust detection system capable of reliably identifying on-ground chestnuts. Such a system must effectively deal with challenges such as occlusion \cite{arakawa2024tanaka}, illumination variations \cite{mccool20162016}, and visually complex backgrounds, where objects such as leaves, stones, and soil share similar color and texture characteristics with chestnuts and can lead to false positives or missed detection. In recent years, YOLO (You Only Look Once) and RT-DETR (Real-Time DEtection TRansformer) models have demonstrated strong performance in agricultural small-target detection tasks. Mamdouh \& Khattab (2021) \cite{Mamdouh2021} employed an improved YOLOv4-based algorithm for olive fruit fly, achieving a precision of 0.84, a recall of 0.97, and a mean Average Precision (mAP) of 96.68\%. Liao et al. (2025) \cite{liao2025yolo} proposed the YOLO-MECD model based on YOLOv11, achieving a precision of 84.4\% and an mAP of 81.6\%. Allmendinger et al. (2025) \cite{allmendinger2025assessing} applied the RT-DETR-l model to weed detection, achieving an average precision of 82.44\% and an average recall of 66.02\%. Mu et al. (2025) \cite{mu2058a} conducted a comparative benchmark of YOLO (v8–v12) and RT-DETR (v1–v2) models for blueberry detection using a large-scale orchard dataset, achieving a maximum mAP@50 of 93.6\% with the RT-DETRv2-X model. Despite these advances, a systematic and comparative evaluation of state-of-the-art real-time detection models for on-ground chestnut detection under real orchard conditions remains largely unexplored.

The overall objective of this study was therefore to systematically evaluate the applicability of state-of-the-art real-time object detection models for on-ground chestnut detection in orchard environments. Specifically, the objectives were to: (1) construct a labeled chestnut detection dataset that reflects real orchard conditions; (2) conduct a comprehensive quantitative comparison of representative real-time object detection models, including YOLOv11, YOLOv12, YOLOv13, as well as RT-DETRv1, RT-DETRv2, RT-DETRv3, and RT-DETRv4, in terms of detection accuracy, robustness under complex field conditions, and real-time inference performance; and (3) analyze the implications of the comparative results for the design and deployment of vision-based real-time automated chestnut harvesting systems.

\section{Materials and Methods} 

\subsection{Chestnut Dataset}

The chestnut dataset used in this study was collected in a commercial orchard (Owosso, Michigan) during the 2024 harvest season. Images were acquired by walking through the orchard while capturing the ground scenes using a hand-held smartphone (iPhone 12). The data collection process encompassed a wide range of ground conditions, including varying grass coverage, soil conditions, and illumination conditions, providing a diverse and representative dataset for robust modeling.

\begin{figure}[htbp]
\centering
\includegraphics[width=0.9\linewidth]{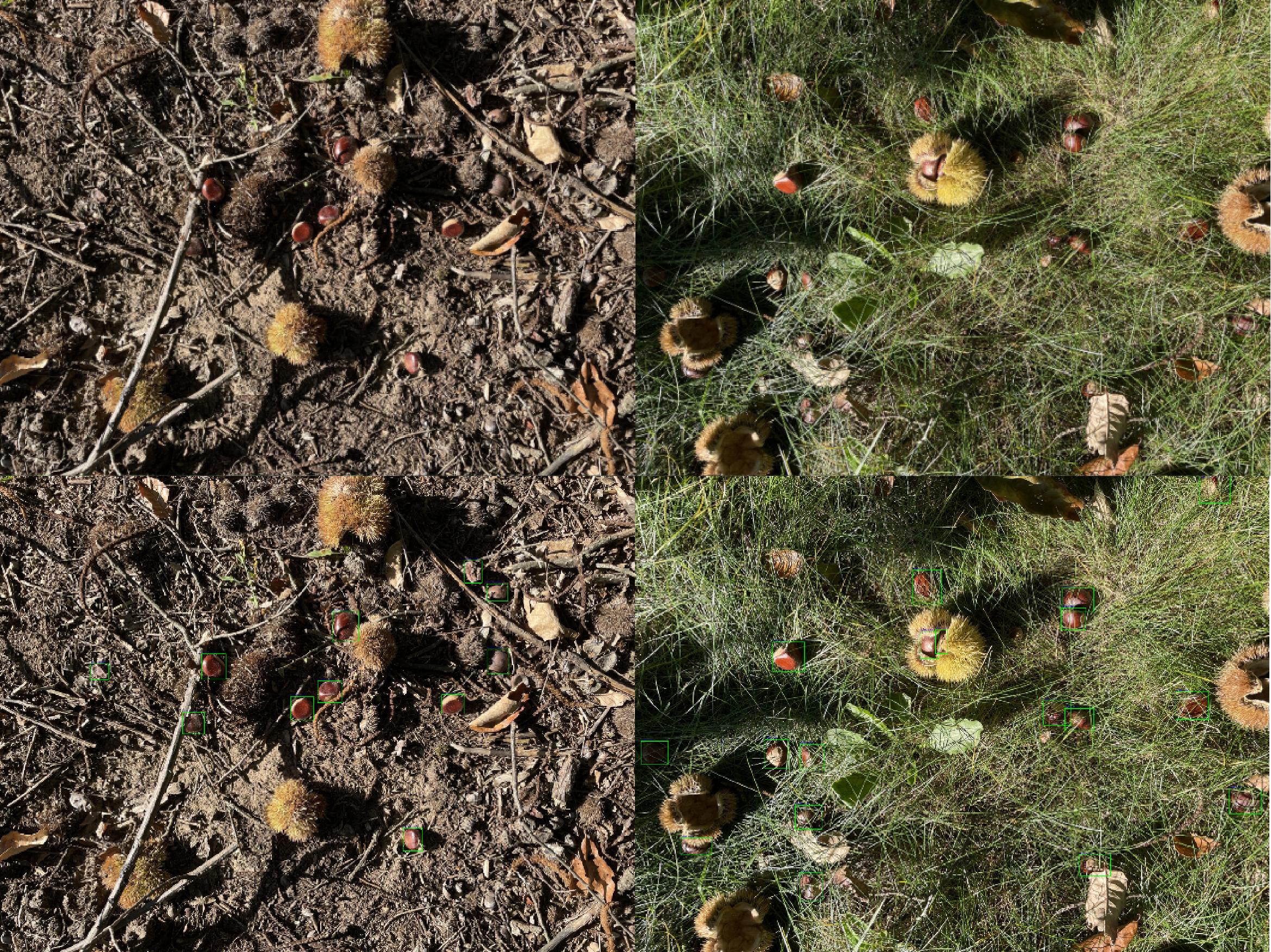}
\caption{Example of original images and annotation images of on-ground chestnuts. The green bounding boxes indicate the labeled chestnuts.}
\label{fig:fig1}
\end{figure}

The acquired images were manually annotated using VGG Image Annotator (VIA) \cite{dutta2019the} by drawing bounding boxes for exposed chestnuts in each image. All images were carefully labeled to capture individual chestnuts, including partially visible and occluded ones. Such meticulous annotation was critical for ensuring model robustness under challenging conditions such as occlusion, shadows, and non-uniform backgrounds. Figure 1 shows representative examples of original images alongside the corresponding images with bounding box annotations. The final annotations were saved in JavaScript Object Notation (JSON) file format, and the consistency and accuracy of the annotations were verified through multiple rounds of review to ensure the quality of the annotations. The resulting dataset consists of 319 high-resolution (4032 × 3024 pixels) color images with a total of 6,524 bounding box annotations. Figure 2 shows the distribution of the number of annotated chestnut instances per image. The chestnut counts per image ranged from 2 to 155, with an average of 23; in particular, approximately 76\% of the dataset contained no more than 30 chestnut instances.

\begin{figure}[htbp]
\centering
\includegraphics[width=0.9\linewidth]{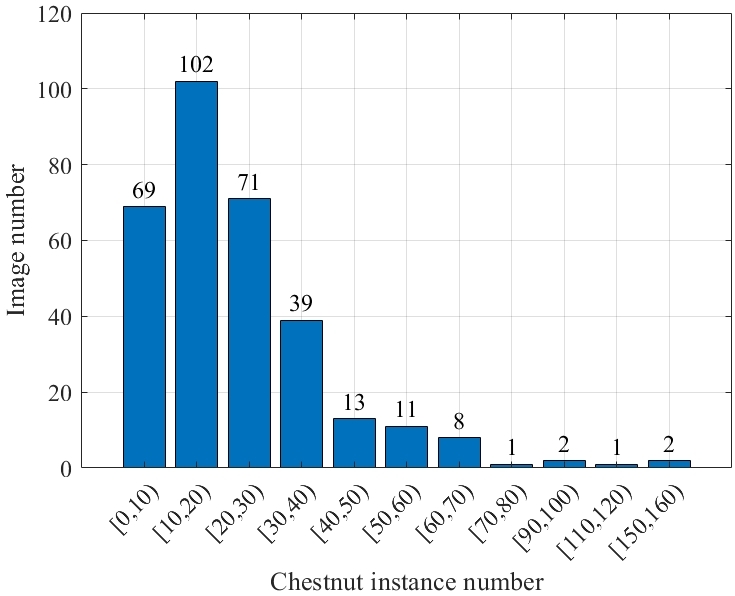}
\caption{The histogram of the annotated chestnuts per image.}
\label{fig:fig2}
\end{figure}

\subsection{Chest Detection Models}

\subsubsection{Real-Time Object Detectors}

Deep-learning-based object detectors are typically categorized into two types: two-stage and one-stage detectors \cite{zhang2021a}. One-stage models in particular offer faster and real-time detection capabilities by eliminating the region proposal step in two-stage detectors and performing detection in a single unified process. For automated chestnut harvesting, rapid and reliable real-time detection is crucial to achieve efficient harvest operations, for which one-stage detectors are more practically appropriate. Among the most widely used real-time object detection frameworks are the YOLO and RT-DETR model families, which are briefly presented below.

\subparagraph{YOLO Series}

YOLO is a real-time object detection framework designed to perform detection in a single forward pass of a neural network. The framework divides the image into a grid and predicts bounding boxes and class probabilities for each cell. The concept of the first version, YOLOv1 \cite{redmon2016you}, is to divide the input image into a grid and each grid cell is given responsibility for detecting if an object's center falls within the grid cell. However, the initial versions struggled with detecting small objects due to coarse grid divisions and relatively lower localization accuracy when compared with region-based methods. Over the past three years, YOLO has undergone substantial architectural and methodological improvements to address these limitations.

YOLOv11, which was introduced in 2024 by Ultralytics \cite{glenn2024ultralytics}, is a comprehensive update to the classic backbone-neck-head paradigm. The incorporation of the Cross-Stage Local Self-Attention (C2PSA) module enables the model to more effectively capture contextual information across multiple layers, thereby improving the accuracy of object detection, especially for small objects. YOLOv12 \cite{tian2025yolov12} introduces the Area Attention (A2) module, which maintains a large receptive field while drastically reducing computational complexity, allowing the model to enhance speed without compromising accuracy. YOLOv13 \cite{lei2025yolov13} addresses the above-mentioned challenges by proposing a Hypergraph-based Adaptive Correlation Enhancement (HyperACE) mechanism. This approach adaptively exploits latent high-order correlations through hypergraph computation, overcoming the limitation of pairwise correlation and enabling efficient global cross-location and cross-scale feature fusion and enhancement.

\subparagraph{RT-DETR Series}

DETR (DEtection TRansformer) represents a paradigm shift in object detection by introducing transformer architectures to model global spatial relationships across an image. Unlike traditional detectors, DETR eliminates the need for anchor boxes and non-maximum suppression, simplifying the detection pipeline. DETR's ability to capture long-range dependencies makes it particularly effective in detecting objects in challenging environments. Recent advancements have improved its convergence speed and performance on small object detection tasks, further expanding its applicability to real-world tasks.

DETR (Detection Transformer) represents a paradigm shift in object detection by introducing transformer architectures to model global spatial relationships across an image. Unlike traditional detectors, DETR eliminates the need for anchor boxes and non-maximum suppression, simplifying the detection pipeline. DETR's ability to capture long-range dependencies makes it particularly effective in detecting objects in challenging environments. Recent advancements have improved its convergence speed and performance on small object detection tasks \cite{zhu2010deformable}, further expanding its applicability to real-world tasks.

RT-DETR (v1) \cite{zhao2024detrs} is the first real-time variant of DETR specifically designed to overcome the high computation cost and low inference speed of the original architecture. It achieves real-time end-to-end object detection by replacing the vanilla Transformer encoder with an efficient hybrid encoder. RT-DETR introduces attention-based intra-scale feature interaction (AIFI) to focus on refining features within the same scale, and CNN-based cross-scale feature fusion to integrate multi-scale information. Additionally, an IoU-Aware query selector is introduced to optimize the decoder's initial target query during the training phase, resulting in higher-quality queries and improved detection performance.

RT-DETRv2 \cite{lv2024rt} enhances the flexibility and practicality optimization of RT-DETR by introducing scale-adaptive sampling in the deformable attention module, enabling features at varying scales to be extracted more efficiently. The model also introduces a mask generator to enhance query diversity through perturbation masks applied in the self-attention step. This reduces redundancies and enhances query-ground truth matching. In terms of optimization, RT-DETRv2 combines L1 loss, generalized IoU (GIoU) loss, and variable focus loss (VFL) to balance regression and classification tasks. These strategies jointly improve detection accuracy and computational efficiency.

RT-DETRv3 \cite{wang2024rt} further improves detection performance by introducing a CNN-based one-to-many label assignment auxiliary head, jointly optimized with the main detection branch to strengthen the encoder's representations. To effectively address the problem of sparse supervision in object detection, RT-DETRv3 proposes a learning strategy with self-attention perturbations to enhance decoder supervision by diversifying label assignments across multiple query groups. Furthermore, it introduces a shared-weight decoder branch for dense forward supervision, ensuring that more high-quality queries are matched to each ground truth. These methods significantly improve model performance and accelerate convergence without increasing inference latency.

Very recently, RT-DETRv4 \cite{liao2025rt} was proposed to advance real-time detection by addressing the representational bottleneck of lightweight architectures while fully preserving inference efficiency. It introduces a training-time semantic distillation framework that leverages the representational power of advanced Vision Foundation Models (VFMs) without modifying the inference-time architecture. To enable stable and task-aligned semantic transfer, a Deep Semantic Injector (DSI) that integrates high-level semantic representations from VFMs is incorporated into the deep layers of the detector, enriching encoder features. In addition, a Gradient-guided Adaptive Modulation (GAM) strategy is designed to dynamically regulate the strength of semantic injection based on gradient norm ratios, harmonizing semantic distillation with detection optimization. These innovations allow RT-DETRv4 to achieve substantial performance gains with no additional inference latency or deployment cost, offering a new state-of-the-art balance between accuracy and speed.

\subsubsection{Model Selection and Configuration}

This study selected YOLOv11, YOLOv12, and YOLOv13, as well as RT-DETRv1, RT-DETRv2, RT-DETRv3, and RT-DETRv4, based on their demonstrated performance across different real-time vision tasks [19, 22, 34], to develop chestnut detection models using the constructed orchard dataset. All detectors were implemented using open-source software packages provided by their respective developers, as summarized in Table 1, and were adapted and retrained for the chestnut detection task in this study. To comprehensively evaluate model performance across capacity levels and to provide deployment flexibility under varying onboard computing resource constraints, multiple architecture variants were considered. Specifically, five variants of YOLOv11 and YOLOv12 (nano, small, medium, large, and extra-large) and four variants of YOLOv13 (nano, small, large, and extra-large) were tested, resulting in a total of 14 YOLO model variants evaluated in this study. Similarly, four backbone variants (R18, R34, R50, and R101) were evaluated for each of RT-DETRv1, RT-DETRv2, and RT-DETRv4, along with three backbone variants (R18, R34, and R50) for RT-DETRv3, yielding a total of 15 RT-DETR model variants.

\begin{table*}[htbp]
\centering
\caption{Summary of YOLO variants and RT-DETR variants with open-source software packages.}
\label{tab:tab1}
\resizebox{\textwidth}{!}{%
\begin{tabular}{l|l|l|l}
\hline
Index & Models & URL & Reference \\
\hline
1 & YOLOv11 & \url{https://github.com/ultralytics/ultralytics} & Jocher \& Qiu (2024) \cite{glenn2024ultralytics}\\
\hline
2 & YOLOv12 & \url{https://github.com/sunsmarterjie/yolov12} & Tian et al. (2025) \cite{tian2025yolov12}\\
\hline
3 & YOLOv13 & \url{https://github.com/iMoonLab/yolov13} & Lei et al. (2025) \cite{lei2025yolov13}\\
\hline
4 & RT-DETRv1 & \url{https://github.com/lyuwenyu/RT-DETR} & Zhao et al. (2023) \cite{zhao2024detrs}\\
\hline
5 & RT-DETRv2 & \url{https://github.com/lyuwenyu/RT-DETR} & Lv et al. (2024) \cite{lv2024rt}\\
\hline
6 & RT-DETRv3 & \url{https://github.com/clxia12/RT-DETRv3} & Wang et al. (2025) \cite{wang2024rt}\\
\hline
7 & RT-DETRv4 & \url{https://github.com/RT-DETRs/RT-DETRv4} & Liao et al. (2025) \cite{liao2025rt}\\
\hline
\end{tabular}
}%
\end{table*}

\subsubsection{Experimentation}

Figure 2 illustrates the overall modeling process for chestnut detection. The original image annotation data (JSON format) provided by the VIA tool was first converted into YOLO-format labels for training YOLOv11, YOLOv12, and YOLOv13. The YOLO format annotations were then converted to the corresponding COCO data format to ensure compatibility with RT-DETR models. Following the format conversion, the dataset was randomly partitioned into three subsets: training, validation, and test subsets, with a split ratio of 70\%, 10\%, and 20\%, corresponding to 223, 31, and 65 images, respectively.

\begin{figure*}[!t]
\centering
\includegraphics[width=1.0\textwidth]{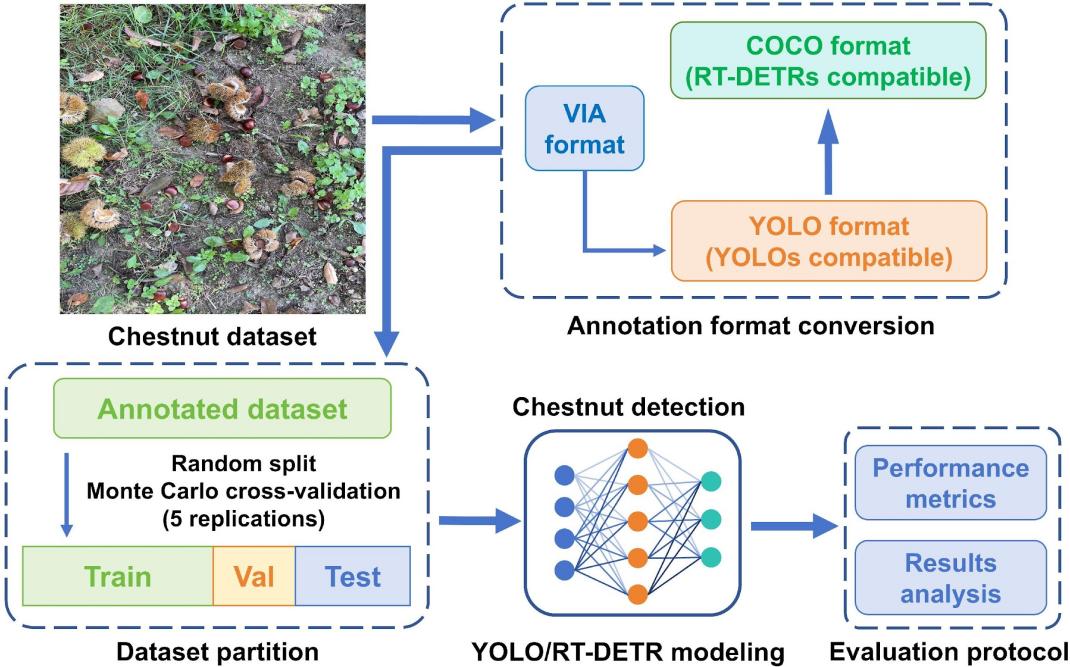}
\caption{The proposed pipeline of chestnut detection by YOLO/RT-DETR object detectors.}
\label{fig:fig3}
\end{figure*}

Model training was conducted using the official implementation released on GitHub, with Python 3.11 on a workstation equipped with eight NVIDIA RTX 4090 GPUs (24 GB each) and a 32-core CPU Intel Xeon Platinum 8481C processor. For the YOLO-based experiments, five variants of YOLOv11 and YOLOv12 (nano, small, medium, large, and extra-large), and four variants of YOLOv13 (nano, small, large, and extra-large) were trained for 200 epochs using the stochastic gradient descent (SGD) optimizer. The initial learning rate was set to 0.01 and adjusted dynamically through cosine learning rate scheduling, with a weight decay value of 0.0005. The training included a warmup phase lasting 3.0 epochs, and the intersection-over-union (IoU) threshold was set at 0.7 for bounding box predictions. The batch size was set at 8, with 12 data-loading workers, and Automatic Mixed Precision (AMP) training was employed to improve computational efficiency and reduce GPU memory consumption.

Since the original dataset images have a high resolution of 4032×3024 pixels, the choice of input size requires balancing detection accuracy and computational efficiency. A larger input size preserves more image details, leading to better performance in small-object detection; however, it also significantly increases computation, slows down training and inference, and consumes more GPU memory. Considering both accuracy requirements and available hardware resources, all images were empirically resized to 1024×1024 pixels. Data augmentation followed the models' default automatic augmentation strategies, such as random flipping and scaling. Model outputs, consisting of confidence scores and bounding boxes, were stored in JSON format for subsequent evaluation. Standard metrics in terms of detection accuracy, model complexity, and inference time (Section 2.3) were employed for model performance assessment, consistent with common evaluation practices in object detection studies \cite{le2025thai}.

For the RT-DETR–based experiments, four backbone variants (R18, R34, R50, and R101) were evaluated for RT-DETRv1, RT-DETRv2, and RT-DETRv4, along with three backbone variants (R18, R34, and R50) for RT-DETRv3. All models were fine-tuned from the official pretrained weights and trained for 200 epochs with the SGD optimizer. The image size and number of workers were kept consistent with the YOLO experiments. The batch size was set to 4, the initial learning rate was set to 0.0005, the weight decay was set to 0.0001, and the maximum gradient clipping norm was set to 0.5, while all other settings followed the default configurations. The same evaluation metrics as described in Section 2.3 were used for performance comparison.

Due to the random nature of dataset partitioning and the small number of samples in a dataset, training a detection model on a specific dataset partition can be sensitive to a single dataset partition, potentially leading to biased model performance; if a particular partitioning happens to be favorable, the results may be overly optimistic. To improve the credibility of model performance evaluation, this study employed Monte Carlo cross-validation (CV) (also known as repeated holdout CV) with five replicates as done in \cite{dang2023yoloweeds}. Specifically, the dataset was randomly split into the three subsets five times using different random seeds, and model performance was evaluated on the test data for each replicate. Final performance metrics were computed as the average across all five replicates (Figure 3). Unlike traditional K-fold cross-validation, which divides the dataset only once, this Monte Carlo approach provides more reliable performance estimates by accounting for variability introduced by different data splits, enhancing the credibility of model performance evaluation.

\subsection{Performance evaluation metrics}

The performance of YOLO and RT-DETR object detectors in chestnut detection was evaluated in terms of detection accuracy, model complexity, and inference times (Dang et al., 2022) \cite{dang2022deepcottonweeds}, which are described below.

\subsubsection{Detection accuracy}

The detection accuracy of trained models was assessed on test data using standard object detection metrics, including precision, recall, mAP@0.5, and mAP@[0.5:0.95]. Among the metrics, the mAP@0.5 is commonly used as the primary indicator of object detection performance, while mAP@[0.5:0.95] provides a more comprehensive assessment across varying localization strictness. The IoU quantifies the overlap between a predicted bounding box and the corresponding ground-truth box and is defined as the ratio of the area of their intersection to the area of their union. In this study, precision and recall were computed using an IoU threshold of 0.7 to reflect stricter localization requirements, whereas mAP@0.5 and mAP@[0.5:0.95] were calculated following the standard COCO protocol with IoU thresholds of 0.5 and 0.5:0.05:0.95.

For a given object class, detection outcomes are defined as follows:

True Positive (TP): a predicted bounding box that correctly matches a ground-truth object of the same class with an IoU greater than or equal to 0.7.

False Positive (FP): a predicted bounding box that either does not correspond to any ground-truth object or is assigned to an incorrect class, or whose IoU with the matched ground-truth object is below the threshold.

False Negative (FN): a ground-truth object that is not detected by the model.

True Negative (TN): background regions correctly identified as non-object areas (typically not explicitly used in object detection metric calculations).

Based on these definitions, precision (P) and recall (R) are computed as:

The average precision (AP) for a given class is defined as the area under the precision–recall (P–R) curve:

The mean average precision (mAP) is then computed as the mean of AP values across all object classes:

\subsubsection{Number of model parameters}

The number of model parameters is a direct measure of model complexity and a key consideration for real-world deployment. Models with a larger number of parameters require more memory during inference and generally incur higher computational costs, leading to increased inference times. Consequently, although larger models may offer improved accuracy, they typically demand greater computational resources and longer inference times, which can limit their suitability for real-time applications.

\subsubsection{Computation cost and inference time}

Floating-point operations (FLOPs) are commonly used to measure the computational complexity of a model, representing the number of arithmetic operations required to process a single input. In this study, computational complexity is reported in terms of GFLOPs (GigaFLOPs), which is a standard unit for large-scale deep learning models. Additionally, inference time, defined as the time required for a trained model to generate predictions for an input image, is a critical metric for real-time detection applications. The inference time for each YOLO and RT-DETR detector was computed as the average prediction time across all images in the test set using the same hardware described above.

\section{Results}

\subsection{Yolo Results}

Figure 4 presents the training curves for mAP@0.5 and mAP@[0.5:0.95] across all evaluation model variants. All architectures exhibited rapid feature learning capabilities during the early training stages. Specifically, within the first 100 epochs, YOLOv11 achieved over 90\% mAP@0.5, whereas YOLOv12 and YOLOv13 reached approximately 75\% and 65\%, respectively. Across all three model series, mAP@[0.5:0.95] exceeded 65\% within the same training window. After approximately 150 training epochs, the detection accuracy across all variants converged and stabilized, indicating that the 200-epoch training schedule adopted was sufficient to achieve stable convergence. This rapid convergence demonstrates the ability of these models to effectively adapt to the dataset for chestnut detection, despite challenging ground-level orchard conditions involving shadows, occlusions, and background clutter.

\begin{figure*}[!t]
\centering
\includegraphics[width=1.0\textwidth]{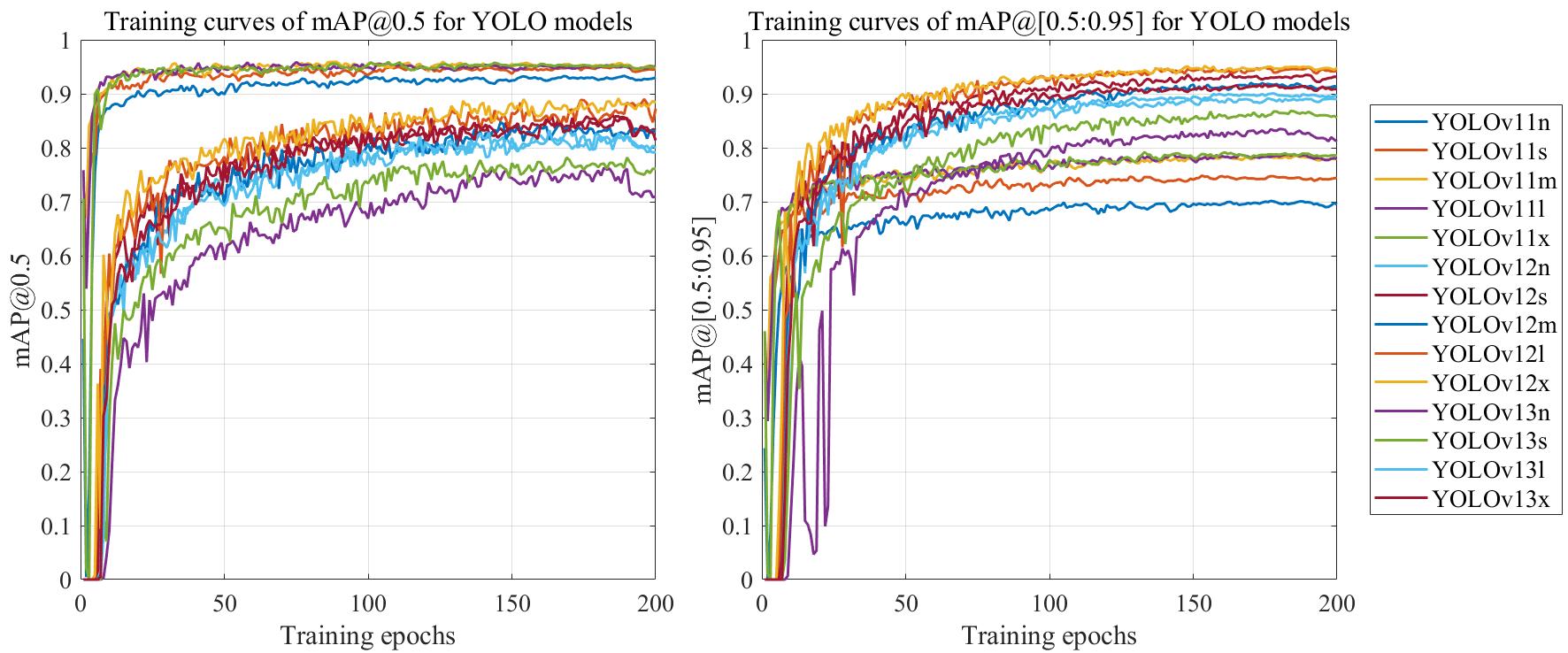}
\caption{Training curves of mAP@0.5 and mAP@[0.5:0.95] for the YOLO models for chestnut detection.}
\label{fig:fig4}
\end{figure*}

Table 2 summarizes the detection performance of all YOLO models on the test dataset. Overall, all three YOLO families achieved competitive performance, with the accuracy generally improving as the model scale increased. Across all variants, mAP@0.5 ranged from 89.8\% (YOLOv13-n) to 95.1\% (YOLOv12-m), while mAP@[0.5:0.95] ranged from 60.5\% (YOLOv13-n) to 80.1\% (YOLOv11-x).

\begin{table*}[htbp]
\centering
\caption{Comparative performance summary of all YOLOv11, v12, and v13 variants. GFLOPs denote giga floating-point operations. ± indicates the standard deviation.}
\label{tab:tab2}
\resizebox{\textwidth}{!}{%
\begin{tabular}{l|l|l|l|l|l|l|l|l}
\hline
Index & Models & Models & Precision (100\%) & Recall (100\%) & mAP@0.5 (100\%) & mAP@[0.5:0.95] (100\%) & GFLOPs & inference time (ms/image) \\
\hline
1 & YOLOv11 & YOLOv11n & 94.8±1.0 & 86.9±2.3 & 92.6±1.4 & 75.0±2.4 & 6.3 & 5.6±0.2 \\
\hline
2 & YOLOv11 & YOLOv11s & 95.5±0.6 & 88.0±1.2 & 93.5±1.4 & 77.5±0.7 & 21.3 & 8.3±0.3 \\
\hline
3 & YOLOv11 & YOLOv11m & 95.5±0.8 & 87.6±1.3 & 93.1±0.6 & 79.1±0.7 & 67.6 & 8.7±0.9 \\
\hline
4 & YOLOv11 & YOLOv11l & 95.3±0.3 & 89.0±0.7 & 93.8±0.5 & 79.9±0.7 & 86.6 & 11.1±2.1 \\
\hline
5 & YOLOv11 & YOLOv11x & 95.3±0.5 & 88.9±1.0 & 93.8±0.5 & \textbf{80.1±0.8} & 194.4 & 16.2±2.7 \\
\hline
 &  & Average & 95.3±0.7 & 88.1±1.6 & 93.4±0.9 & 78.4±2.3 & 75.2 & 10.0±1.2 \\
\hline
6 & YOLOv12 & YOLOv12n & 91.3±1.0 & 83.7±1.2 & 91.6±0.8 & 65.2±1.2 & 6.3 & 8.3±0.5 \\
\hline
7 & YOLOv12 & YOLOv12s & 92.7±0.7 & 86.3±1.2 & 94.0±0.6 & 70.5±0.8 & 21.2 & 11.7±0.7 \\
\hline
8 & YOLOv12 & YOLOv12m & 92.9±1.0 & 89.3±1.8 & 95.1±1.1 & 73.8±2.4 & 67.1 & 11.7±1.5 \\
\hline
9 & YOLOv12 & YOLOv12l & 92.9±0.9 & 88.1±1.2 & \textbf{94.9±0.8} & 74.3±1.9 & 88.5 & 14.3±1.7 \\
\hline
10 & YOLOv12 & YOLOv12x & 92.4±2.0 & 88.6±1.1 & 94.6±0.5 & 73.8±0.7 & 198.5 & 22.1±2.1 \\
\hline
 &  & Average & 92.8±1.1 & 87.6±1.8 & 94.5±1.1 & 71.5±2.4 & 76.3 & 13.6±1.3 \\
\hline
11 & YOLOv13 & YOLOv13n & 90.1±0.4 & 80.5±1.2 & 89.8±0.6 & 60.5±1.4 & 6.2 & 11.9±1.0 \\
\hline
12 & YOLOv13 & YOLOv13s & 92.1±0.6 & 84.0±0.9 & 92.3±0.6 & 66.4±1.2 & 20.7 & 14.8±1.5 \\
\hline
13 & YOLOv13 & YOLOv13l & 91.4±1.9 & 83.7±3.3 & 92.0±2.3 & 66.3±4.6 & 88.1 & 30.7±2.3 \\
\hline
14 & YOLOv13 & YOLOv13x & 91.2±3.3 & 83.0±6.0 & 91.0±4.7 & 66.2±8.6 & 198.7 & 47.8±4,2 \\
\hline
 &  & Average & 91.2±2.8 & 82.8±3.9 & 91.8±2.8 & 64.9±5.6 & 78.4 & 26.3±2.3 \\
\hline
\end{tabular}
}%
\end{table*}

The YOLOv12-m model achieved the highest mAP@0.5 value of 95.1\% and the highest recall value of 89.3\%, while maintaining a high precision value of 92.9\%. These results demonstrate that YOLOv12-m has strong detection accuracy and achieves a good balance between precision and recall. Figure 5 further illustrates an example image of the detection results of YOLOv12-m under complex lighting and severe occlusion conditions. In this test, the model correctly detected 47 out of 49 chestnuts, with only one false positive, achieving a precision of 97.9\% and a recall of 95.9\%. In contrast, YOLOv11-x attained the highest mAP@[0.5:0.95] (80.1\%) and a high precision of 95.3\%, with recall=88.9\%, suggesting superior bounding-box localization accuracy and robustness across varying IoU thresholds, which are important for reducing false positives in practical applications.

Among the three model families, the YOLOv11 model consistently performed well in terms of precision and recall, with all variants achieving a precision exceeding 94\% and a mean precision of 95.3\%. Compared to YOLOv11, YOLOv12 showed a slight decrease in mean precision, but its recall remained similar (87.6\%). Notably, YOLOv12 exhibited significant improvements in recall for medium, large, and super-large variants, indicating that its attention-based architecture, even with the introduction of some false positives, recovered more true positive results. Although YOLOv12 slightly outperformed YOLOv11 in terms of mAP@0.5, its mAP@[0.5:0.95] metric (71.5\%) was significantly lower than YOLOv11's (78.4\%), indicating weaker performance under stricter localization criteria. In contrast, YOLOv13 performs poorly across all scales, particularly on the mAP@[0.5:0.95] scale, where its average score is only 64.86\%. Even its best-performing variant (YOLOv13-s: precision = 92.1\%, recall = 84.0\%, mAP@0.5 = 92.3\%, mAP@[0.5:0.95] = 66.4\%) lags behind YOLOv11 and YOLOv12. This suggests that YOLOv13 may require alternative training strategies, such as lowering the initial learning rate, to more effectively learn finer-grained details and achieve more stable convergence.

\begin{figure}[htbp]
\centering
\includegraphics[width=0.9\linewidth]{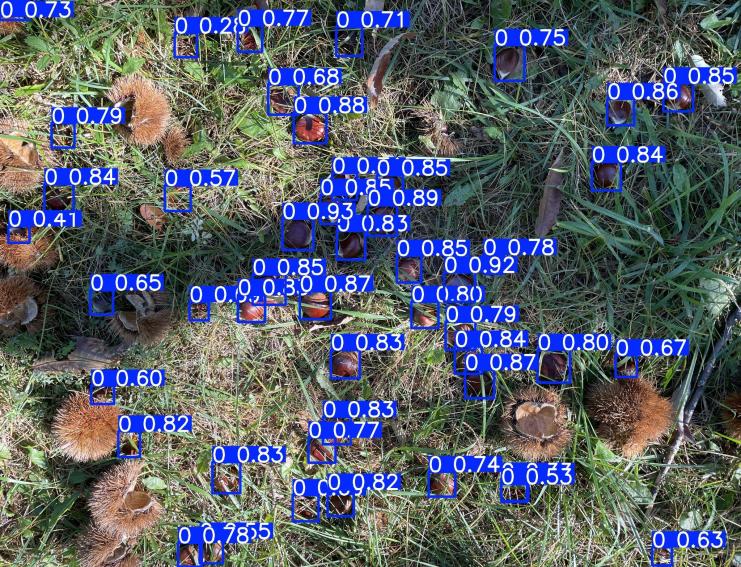}
\caption{Test image of YOLOv12-m under complex lighting and severe occlusion conditions.}
\label{fig:fig5}
\end{figure}

Figure 6 illustrates the relationship between model complexity and computational performance, showing the trends of GLOPs and inference time versus the number of model parameters. As the model scale increases, both inference time and GFLOPs exhibit an upward trend. Across corresponding scales, the three YOLO families demonstrated comparable computational complexity, with YOLOv11-x, YOLOv12-x, and YOLOv13-x all approaching 200 GFLOPs. Although these larger models incurred higher computational costs, their inference time remained below 50 ms, corresponding to frame rates exceeding 20 FPS. This performance still meets the basic real-time detection requirements of ground-based chestnut harvesting systems.

YOLOv11 models appeared to exhibit the most favorable balance between accuracy and computational efficiency. YOLOv11-n achieved the fastest inference time of 5.6 ms, followed by YOLOv11-s at 8.3 ms, which—combined with its precision of 95.5\% and mAP@0.5 of 93.45\%—makes it particularly attractive for embedded and edge device-based applications. YOLOv12-m achieves an inference time of 11.7 ms, while delivering the highest mAP@0.5 of 95.1\%, supporting its feasibility for real-time deployment. In contrast, YOLOv13-x exhibited the highest computational load (198.7 GFLOPs) and longest inference (47.8 ms), rendering it the least suitable for real-time deployment.

\begin{figure*}[!t]
\centering
\includegraphics[width=1.0\textwidth]{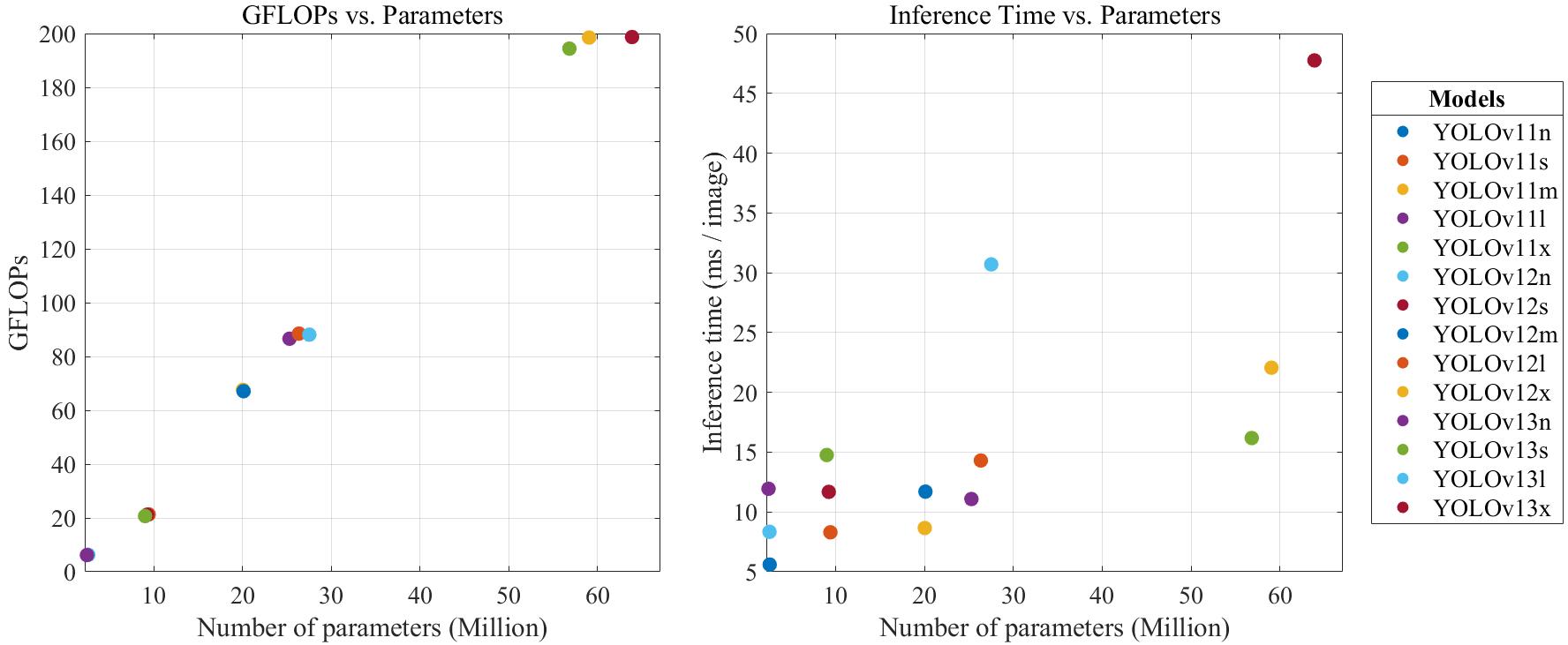}
\caption{Relationship between number of parameters and (a) GFLOPs, (b) inference time for all YOLOv11–YOLOv13 variants. GFLOPs denote giga floating-point operations.}
\label{fig:fig6}
\end{figure*}

\subsection{RT-DETR Results}

Figure 7 presents the training curves of mAP@0.5 and mAP@[0.5:0.95] for all the RT-DETR (v1-v4) models. All the model variants achieved mAP@0.5 values exceeding 80\% and mAP@[0.5:0.95] values above 65\% with the first 60 training epochs, followed by performance stabilization after approximately 100 epochs. These trends suggested effective learning and convergence across all models and confirmed that the 200-epoch training schedule adopted in this study was sufficient to achieve stable convergence.

\begin{figure*}[!t]
\centering
\includegraphics[width=1.0\textwidth]{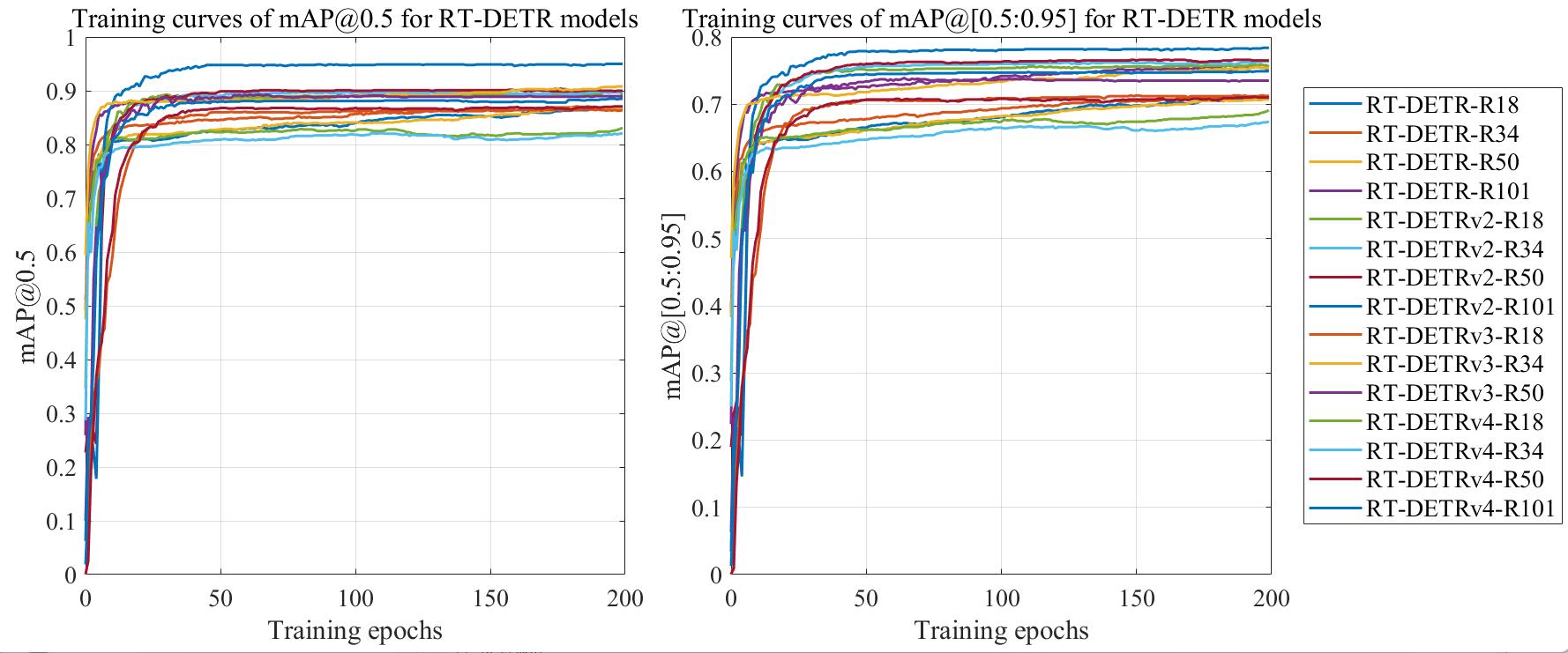}
\caption{Training curves of mAP@0.5 and mAP@[0.5:0.95] for the RT-DERTv1, v2, v3, and v4 models for chestnut detection.}
\label{fig:fig7}
\end{figure*}

Table 3 summarizes the detection performance of the evaluated RT-DETR variants. Similar to the YOLO results, detection accuracy generally improved with increasing model scale. RT-DETRv2 demonstrated substantial improvements over RT-DETRv1 across all evaluation metrics, indicating the effectiveness of its architectural refinements. In contrast, RT-DETRv3 exhibited slightly lower precision and recall than v1 than RT-DETRv1, although it achieved marginal improvements in mAP metrics. Under the same training conditions and using the same dataset, RT-DETRv4 performs worse than previous variants. While RT-DETRv4 aims to leverage a VFM-based semantic distillation framework, which enriches feature representations by injecting high-level semantics from a large pre-trained model into the detector during training, this increased training complexity does not necessarily translate into improved accuracy for tasks where datasets emphasize fine local details. Specifically, RT-DETRv4's distillation mechanism focuses on aligning the backbone network representation with that of a large vision foundation model to enhance overall semantic feature quality. Although this strategy contributes to richer semantic representations and has been shown to improve performance on large, diverse benchmarks such as COCO, it can reduce the model's sensitivity to fine-grained spatial cues when applied to tasks dominated by small, densely packed, and partially occluded objects. In such scenarios, the additional semantic supervision may shift learning emphasis away from the precise localization features required for small object detection, particularly in specialized datasets with limited training diversity, thereby resulting in inferior performance compared with earlier RT-DETR variants.

Among all models, RT-DETRv2-R101 achieved the best overall performance (precision = 95.1\%, recall = 86.3\%, mAP@0.5 = 91.1\%, mAP@[0.5:0.95] = 71.9\%). The precision and recall are comparable to those of the best-performing YOLOv11-series models, highlighting its strong detection capability.

RT-DETRv4 is designed to leverage a VFM-based semantic distillation framework, which enriches feature representations by injecting high-level semantics from a large pre-trained model into the detector during training. However, this increased training complexity does not necessarily translate into improved accuracy for tasks that emphasize fine-grained local details. Specifically, the distillation mechanism in RT-DETRv4 focuses on aligning the backbone network representations with those of a large VFM to enhance overall semantic quality. While this strategy has been shown to improve performance on large, diverse benchmarks such as COCO, it can reduce sensitivity to fine spatial cues when applied to datasets dominated by small, densely packed, and partially occluded objects, such as the chestnut dataset in this study. In such scenarios, the additional semantic supervision may shift learning emphasis away from precise localization features that are critical for small object detection, particularly in specialized datasets with limited training diversity. This likely explains the inferior performance compared with earlier RT-DETR variants.

\begin{table*}[htbp]
\centering
\caption{Comparative performance summary of RT-DERTv1, v2, v3, and v4 variants. GFLOPs denote giga floating-point operations. ± indicates the standard deviation.}
\label{tab:tab3}
\resizebox{\textwidth}{!}{%
\begin{tabular}{l|l|l|l|l|l|l|l|l}
\hline
Index & Models & Models & Precision (100\%) & Recall (100\%) & mAP@0.5 (100\%) & mAP@[0.5:0.95] (100\%) & GFLOPs & inference time (ms/image) \\
\hline
1 & RT-DETRv1 & RT-DETRv1-R18 & 88.5±3.2 & 81.2±2.8 & 84.3±0.6 & 68.5±0.5 & 152 & 72.1±3.7 \\
\hline
2 & RT-DETRv1 & RT-DETRv1-R34 & 90.7±2.5 & 82.6±3.2 & 86.8±0.6 & 70.6±0.4 & 233 & 81.3±2.5 \\
\hline
3 & RT-DETRv1 & RT-DETRv1-R50 & 92.1±2.4 & 82.9±2.0 & 87.7±0.5 & 73.2±0.7 & 342 & 90.4±4.3 \\
\hline
4 & RT-DETRv1 & RT-DETRv1-R101 & 92.8±2.6 & 83.1±2.4 & 89.3±0.8 & 74.5±0.6 & 656 & 103.5±4.5 \\
\hline
 &  & Average & 91.0±2.7 & 82.5±2.6 & 87.0±0.6 & 71.7±0.6 & 346 & 86.8±3.7 \\
\hline
5 & RT-DETRv2 & RT-DETRv2-R18 & 92.5±1.8 & 80.1±4.0 & 88.3±1.4 & 73.7±0.8 & 153 & 47.4±1.5 \\
\hline
6 & RT-DETRv2 & RT-DETRv2-R34 & 94.2±2.2 & 82.9±3.2 & 89.4±1.1 & 75.1±1.2 & 234 & 48.8±2.4 \\
\hline
7 & RT-DETRv2 & RT-DETRv2-R50 & 94.7±2.6 & 84.7±3.0 & 90.0±1.8 & 70.5±1.1 & 343 & 56.1±2.3 \\
\hline
8 & RT-DETRv2 & RT-DETRv2-R101 & 95.1±1.5 & 86.3±2.4 & \textbf{91.1±1.6} & 71.9±0.8 & 657 & 66.3±3.2 \\
\hline
 &  & Average & 94.1±2.0 & 83.5±3.2 & 89.7±1.5 & 72.8±1.0 & 347 & 54.6±2.4 \\
\hline
9 & RT-DETRv3 & RT-DETRv3-R18 & 87.6±3.2 & 79.4±1.8 & 84.5±3.2 & 67.2±2.7 & 154 & 52.7±2.4 \\
\hline
10 & RT-DETRv3 & RT-DETRv3-R34 & 90.6±3.4 & 82.5±2.5 & 87.8±2.6 & 71.8±2.6 & 235 & 59.2±2.5 \\
\hline
11 & RT-DETRv3 & RT-DETRv3-R50 & 92.7±3.8 & 84.5±3.4 & 90.4±3.0 & \textbf{78.9±2.5} & 344 & 67.1±4.7 \\
\hline
 &  & Average & 90.3±3.5 & 82.1±2.6 & 87.6±3.0 & 72.6±2.6 & 244 & 59.7±3.2 \\
\hline
12 & RT-DETRv4 & RT-DETRv4-R18 & 85.6±1.8 & 80.3±2.7 & 80.3±1.8 & 63.6±2.5 & 78 & 23.7±2.1 \\
\hline
13 & RT-DETRv4 & RT-DETRv4-R34 & 87.0±2.0 & 82.6±2.5 & 83.5±3.4 & 66.5±2.6 & 167 & 27.3±2.6 \\
\hline
14 & RT-DETRv4 & RT-DETRv4-R50 & 88.4±1.4 & 83.9±1.8 & 84.6±3.2 & 70.9±2.7 & 286 & 30.5±3.3 \\
\hline
15 & RT-DETRv4 & RT-DETRv4-R101 & 89.5±2.1 & 85.5±2.4 & 86.1±2.9 & 72.0±2.0 & 658 & 38.0±3.7 \\
\hline
 &  & Average & 87.6±1.8 & 83.1±2.4 & 83.7±2.8 & 68.3±1.2 & 297 & 29.89±2.88 \\
\hline
\end{tabular}
}%
\end{table*}

Figure 8 illustrates the relationship among the model size (number of parameters), computational complexity (GFLOPS), and inference time for all evaluated RT-DETR models. Consistent with the trends observed in the YOLO experiments, both inference time and computational cost generally increase with model size. Compared to RT-DETRv1, v2, and v3, the RT-DETRv4 architecture substantially reduces the number of parameters across all backbone scales. Specifically, RT-DETRv4-R18 contains only 10 million parameters, representing a 50\% reduction relative to the 20 million parameters used in corresponding R18 variants of v1-v3. At even larger scales, RT-DETRv4-R34 and RT-DETRv4-R50 reduce parameter counts from 31 million and 42 million to 19 million and 31 million, respectively. Even at its largest scale, RT-DETRv4-R101 uses only 62 million parameters, approximately 18\% less than the 76 million parameters required by the v1 to v3 versions. These reductions in model size translated directly to improved inference efficiency.

Among all RT-DETR variants, RT-DETRv4-R18 achieved the fastest inference time (23.7 ms), substantially outperforming other models using the same backbone network, including RT-DETRv1-R18 (72.1 ms), RT-DETRv2-R18 (47.4 ms), and RT-DETRv3-R18 (52.7 ms). In contrast, the model achieving the highest mAP@0.5, RT-DETRv2-R101, exhibited an inference time of 66.3 ms. Although all RT-DETR models satisfied basic real-time requirements, their overall inference speed remained slower than those of the YOLO-based detectors evaluated in this study.

\begin{figure*}[!t]
\centering
\includegraphics[width=1.0\textwidth]{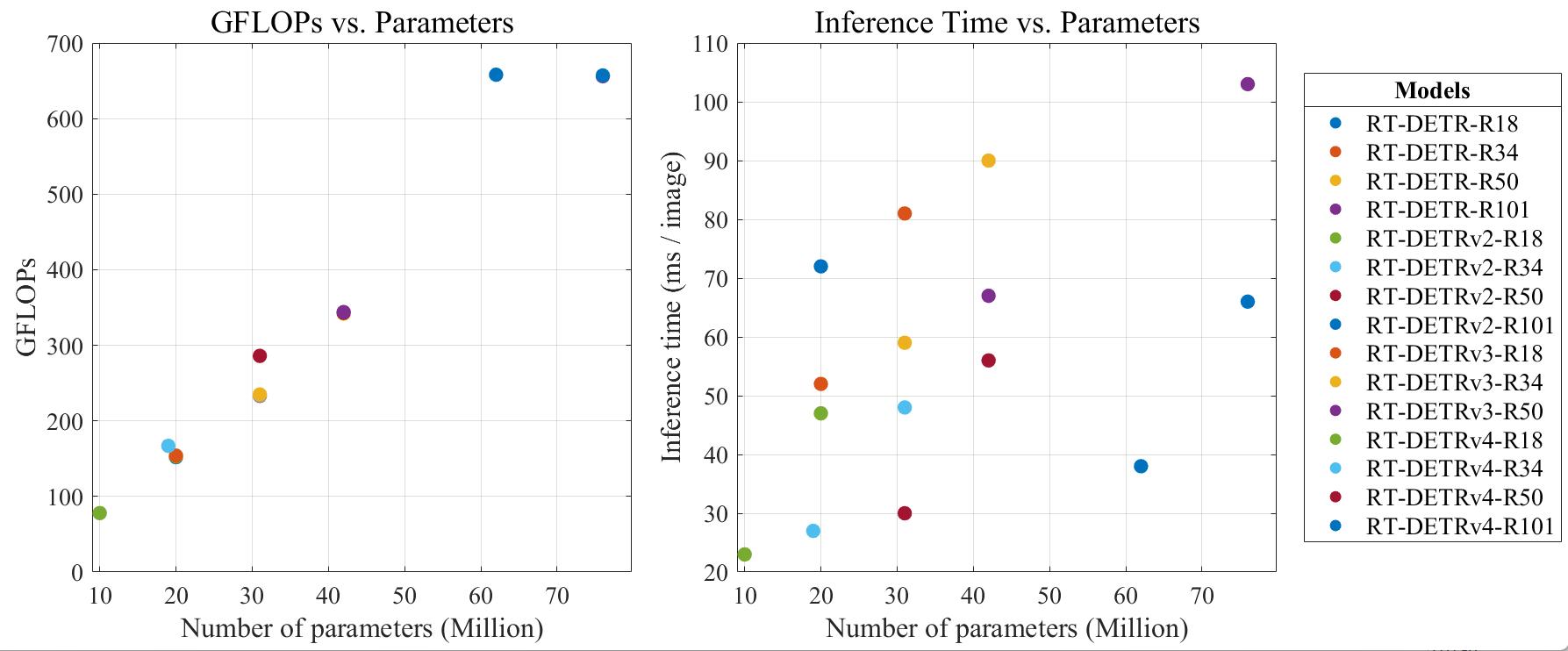}
\caption{Relationship between number of parameters and (a) GFLOPs, (b) inference time for all RT-DETR (v1-v4) variants. GFLOPs denote giga floating-point operations.}
\label{fig:fig8}
\end{figure*}

\subsection{YOLO vs RT-DETR}

Figure 9 illustrates the trade-off between detection accuracy (mAP@0.5 and mAP@[0.5:0.95]) and inference time for the evaluated YOLO and RT-DETR models. As noted above, the RT-DETR models, especially v1–v3, exhibited substantially slower inference speeds than the YOLO-based detectors. Although RT-DETR variants achieved mAP@0.5 values above 85\%, their overall performance across both mAP metrics remained inferior to that of the best-performing YOLO models. When inference time and precision are jointly considered, the YOLOv11 family demonstrates the most favorable balance for real-time chestnut detection, providing consistently high accuracy with significantly lower latency, and thus representing the most practical choice for deployment in harvesting systems. Figure 10 presents representative detection results from RT-DETRv2-R101, the strongest RT-DETR variant, applied to the same test image shown in Figure 5. In this case, RT-DETRv2-R101 detected 44 out of 49 chestnuts with a precision of 95.1\%, which is lower than the 97.9\% achieved by YOLOv12m under identical conditions.

\begin{figure*}[!t]
\centering
\includegraphics[width=1.0\textwidth]{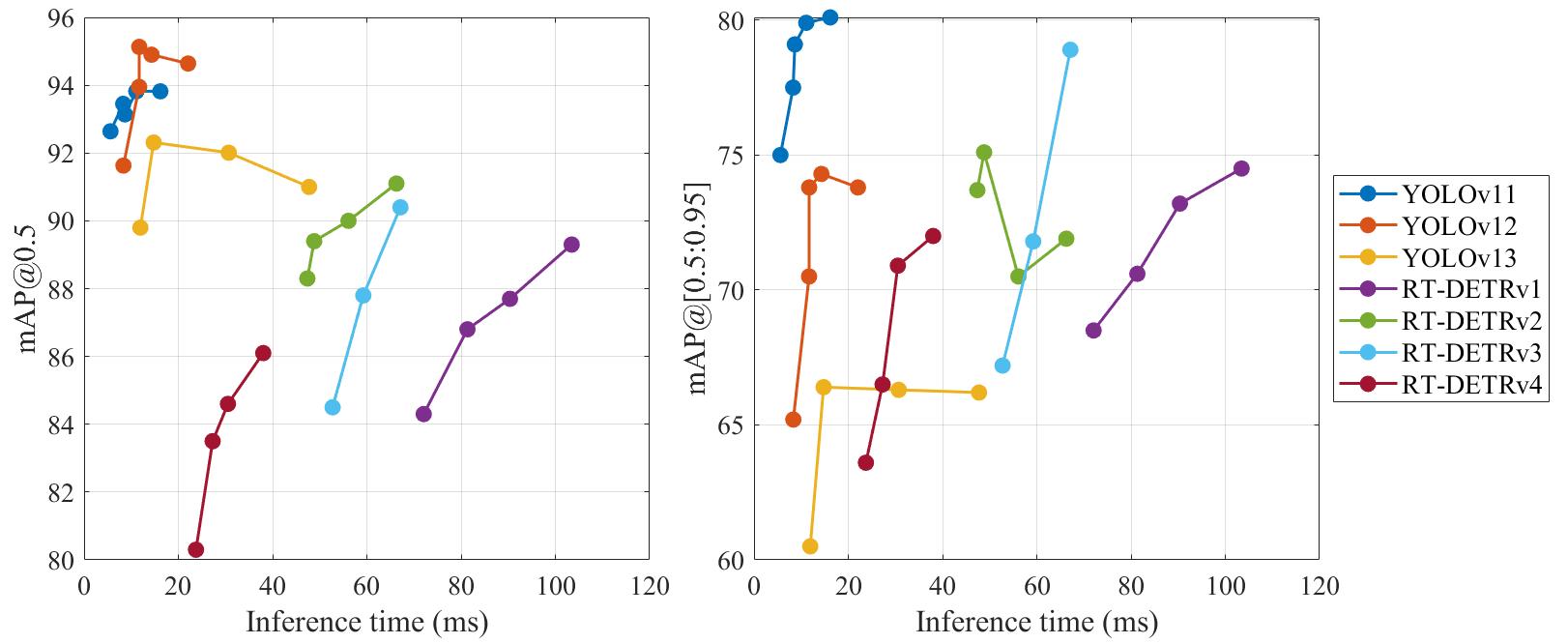}
\caption{Overall comparison between YOLO and RT-DETR variants for chestnut detection.}
\label{fig:fig9}
\end{figure*}

\begin{figure}[htbp]
\centering
\includegraphics[width=0.9\linewidth]{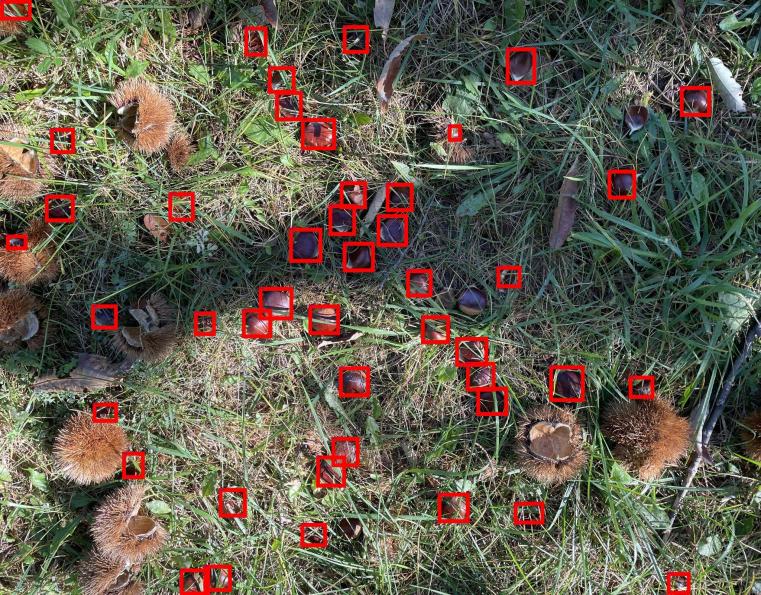}
\caption{Test image of RT-DETRv2-R101 under complex lighting and severe occlusion conditions.}
\label{fig:fig10}
\end{figure}

Across multiple evaluation metrics, the performance gap between RT-DETR and YOLOv11/YOLOv12 indicates architectural differences between the two frameworks. RT-DETR models are designed to capture global contextual information and long-range spatial dependencies, which can be advantageous for complex scene understanding. However, this design can limit their ability to preserve the fine-grained spatial features required for the precise localization of small, densely distributed, and partially occluded chestnuts. In contrast, YOLO architectures emphasize hierarchical local feature extraction and multi-scale feature fusion, enabling more robust performance under the complex, ground-level orchard environments. Overall, these results suggest that while RT-DETR may be advantageous for certain complex tasks, YOLO models are more appropriate for chestnut detection applications that demand both high accuracy and real-time processing capability.

\section{Discussion}

Due to the complexity of orchard floor environments, research on ground-level chestnut detection in commercial orchard conditions remains limited. The results of this study demonstrate that both the emerging YOLO and RT-DETR models can effectively identify chestnuts in realistic orchard settings. Within the YOLO family, YOLOv11 consistently achieved the best overall detection performance, outperforming YOLOv12 and YOLOv13. These findings are consistent with those reported by Sapkota et al. (2024) \cite{sapkota2040comprehensive}, who evaluated multiple YOLO variants (v8-v12) for in-orchard pre-sparse detection of green apples and found that YOLOv11 demonstrated excellent precision, while YOLOv12-1 achieved the highest recall. Together,  these results reinforce the robustness and suitability of YOLO models in chestnut detection. YOLOv11's architecture preserves fine spatial details, enabling tight bounding box localization, while its relatively small model size also supports fast inference speed, making it promising for embedded deployment on harvesting equipment.

In contrast, RT-DETR models exhibited significantly longer inference time than that of the YOLO series models. This observation aligns with findings from Saltık et al. (2024) \cite{salt2024comparative}, who reported that RT-DETRv1 can achieve competitive mAP at larger image sizes, but at the expense of increased inference time. These characteristics suggest that RT-DETR may be better suited for offline or batch processing scenarios, where global contextual modeling is beneficial and real-time constraints are less stringent.

Several limitations of this study warrant further investigation. First, the chestnut dataset is relatively small, which may limit the generalizability of the findings to orchards with different cultivars, soil types, ground terrains, or harvesting conditions. Second, although both CNN-based YOLO models and Transformer-based RT-DETR models were evaluated, the overall training configuration—including data augmentation strategies and hyperparameter settings—was primarily developed based on the YOLO family. YOLO detectors are well-suited to aggressive geometric and photometric data augmentation, whereas RT-DETR models, due to their query-based Transformer architecture, are generally more sensitive to augmentation strength, query configurations, and training schedules. While RT-DETR was also tuned through adjustments of key hyperparameters such as learning rate and training schedule, further refinement of augmentation pipelines and other model-specific training strategies may still be required to fully exploit its potential performance. The modeling experiments were conducted using static images; real-world deployment will require validation under continuous video streams, varying illumination throughout the day, and mechanical vibrations from harvesting equipment. Motion blur has been shown to degrade image information acquisition and negatively affect object detection tasks in precision agriculture scenarios \cite{deng2025development}, indicating the need for models robust against motion for reliable field performance. Moreover, future work will focus on expanding the scale and diversity of the dataset, further optimizing RT-DETR-specific training configurations, and validating the performance of models deployed on harvesting platforms in dynamic conditions. This study focused exclusively on chestnut detection and did not explore downstream tasks such as vision-mechanisms integration, vision-guided chestnut picking, and harvesting platform locomotion, which are necessary for developing an autonomous chestnut harvesting system.

From a model-design perspective, the observed performance differences among YOLO variants are closely related to their ability to preserve and exploit high-resolution features for small-object representation under complex backgrounds. Ground-level chestnuts are typically small, densely distributed, and frequently occluded by grass, leaves, or soil, which makes the retention of shallow spatial details and local structural cues particularly critical. Excessive spatial downsampling in deeper network layers can suppress weak target responses, leading to missed detections or imprecise localization. Models that more effectively leverage multi-scale feature fusion, therefore tend to achieve higher recall and better localization consistency, particularly under stricter IoU thresholds where accurate boundary regression is essential. This observation suggests that fine-grained spatial information plays a dominant role in distinguishing chestnuts from visually similar background elements.

These findings further indicate that future improvements should prioritize architectural adaptations that explicitly enhance small-object sensitivity, such as strengthening the propagation of high-resolution features throughout the backbone–neck pathway or introducing additional fine-scale detection heads to better capture subtle object cues. Optimizing fusion strategies to balance shallow and deep representations may help reduce the dominance of heavily downsampled features, thereby mitigating the loss of boundary, contour, and texture information that is critical for reliable discrimination in cluttered scenes. Moreover, for different YOLO detection heads, task-specific calibration of scale representation and matching mechanisms may be necessary to ensure stable performance in dense and partially occluded scenarios, where overlapping targets and ambiguous background patterns can otherwise lead to inconsistent learning behavior and degraded detection accuracy.

Research is ongoing to develop a vision-guided chestnut-picking robot by integrating the detection model with a robotic arm and harvesting end-effector, forming a closed-loop system spanning 2D detection, 3D positioning, and harvesting execution. Vision-based fruit recognition and 3D reconstruction remain major challenges in agricultural robotics, especially in outdoor orchard environments with varying lighting, occlusion, and clutter. Prior studies have demonstrated that combining 2D detection with depth or stereo information can achieve accurate and efficient 3D localization. For example, Zhou et al. (2024) \cite{zhou20243d} replaced the complex 3D CNN with a lightweight 2D and stereo vision, while Ge et al. (2023) \cite{ge2023three} showed that 2D bounding box-based depth estimation was both faster and more accurate than full 3D clustering approaches. These methods provide valuable references for chestnut localization in ground environments. From a systems perspective, successful integration will require seamless coordination between vision modules, robotic arms, end effectors, and control units, as emphasized by Chen et al. (2024) \cite{chen2024key}. Given the predominance of small-scale chestnut production in the U.S., future harvesting systems must balance high performance with cost-effectiveness. Vision-guided, AI-powered robotics offers a promising pathway toward flexible, lightweight, cost-effective harvesting automation solutions for small and mid-scale producers.

\section{Conclusions}

Ground chestnut detection is crucial for enabling vision-based automated chestnut picking. In this study, a new dataset of 319 ground-level images acquired in commercial orchard environments was created, comprising a total of 6,524 annotated chestnuts. The state-of-the-art real-time object detectors, including YOLO (v11-v13) and RT-DETR (v1-v4) families, were systematically evaluated for ground chestnut detection. Experiments across multiple model scales demonstrated that YOLOv11 consistently achieved superior localization accuracy at stricter IoU thresholds, achieving a maximum mAP@[0.5:0.95] of 80.1\%. Among all evaluated models, YOLOv11s provided the most favorable balance between detection accuracy and inference speed, making it well-suited for real-time, embedded deployment. YOLOv12m achieved the highest mAP@0.5 of 95.1\% and recall of 89.3\%, highlighting its effectiveness, while YOLOv13 showed no clear performance advantage on this dataset. Although RT-DETR models demonstrated competitive performance, they were overall inferior to YOLO detectors in both detection accuracy and efficiency. Future work will focus on integrating the high-performing detection models into a fully automated chestnut picking system and validating performance under more diverse orchard conditions. The methods and insights from this study are broadly applicable to other small-target crop detection tasks and support the advancement of vision-based intelligent agricultural automation.

\section*{Author Contributions}
K.F.: writing — original draft, investigation, formal analysis, software; Y.L.: writing—original draft, review and editing, conceptualization, data curation, investigation, supervision; X.M.: data curation, investigation. 

\section*{Data Availability Statement}
The chestnut dataset and AI models developed in this study are publicly accessible at \url{https://github.com/AgFood-Sensing-and-Intelligence-Lab/ChestnutDetection}.

\typeout{}
\bibliography{Chestnut_preprint}

@misc{census2022available,
  title = {2022 Census of Agriculture 2022 Census of Agriculture | National Agricultural Statistics Service},
  url = {https://www.nass.usda.gov/Publications/AgCensus/2022/index.php},
  note = {accessed on 1 February 2026}
}

@article{kang2008development,
  author = {Kang, W. S. and Guyer, D.},
  title = {Development of Chestnut Harvesters for Small Farms},
  journal = {Journal of Biosystems Engineering},
  year = {2008},
  volume = {33},
  number = {6},
  pages = {384-389}
}

@article{massantini2021evaluating,
  author = {Massantini, R. and Moscetti, R. and Frangipane, M. T.},
  title = {Evaluating progress of chestnut quality: A review of recent developments},
  journal = {Trends in Food Science \& Technology},
  year = {2021},
  volume = {113},
  pages = {245-254}
}

@article{jermini2006influence,
  author = {Jermini, M. and Conedera, M. and Sieber, T. N.},
  title = {Influence of fruit treatments on perishability during cold storage of sweet chestnuts},
  journal = {Journal of the Science of Food and Agriculture},
  year = {2006},
  volume = {86},
  number = {6},
  pages = {877-885}
}

@article{lee2016efficacy,
  author = {Lee, U. K. and Joo, S. and Klopfenstein, N. B.},
  title = {Efficacy of washing treatments in the reduction of post-harvest decay of chestnuts (Castanea crenata ‘Tsukuba’) during storage},
  journal = {Canadian Journal of Plant Science},
  year = {2016},
  volume = {96},
  number = {1},
  pages = {1-5}
}

@article{sieber2007effects,
  author = {Sieber, T. N. and Jermini, M. and Conedera, M.},
  title = {Effects of the harvest method on the infestation of chestnuts (Castanea sativa) by insects and moulds},
  journal = {Journal of Phytopathology},
  year = {2007},
  volume = {155},
  number = {7‐8},
  pages = {497-504}
}

@misc{guyer20122012,
note = {Guyer, Daniel \& Donis-Gonzalez, Irwin \& Burns, James \& DeKleine, Mark. (2012). Is internal quality of chestnuts influenced by harvest methods and physical stresses? Acta Horticulturae. 1019. 10.17660/ActaHortic.2014.1019.18. doi:10.17660/ActaHortic.2014.1019.18}
}

@misc{colantoni2014carletti,
note = {Colantoni, Andrea \& Moscetti, Roberto \& L., Carletti \& Cecchini, Massimo \& Monarca, Danilo \& Stella, Elisabetta \& G., Menghini \& Speranza, Stefano \& Massantini, Riccardo \& M., Contini \& Manzo, Alberto. (2014). Quality maintenance and storability of chestnuts manually and mechanically harvested. Acta horticulturae. 1043. 10.17660/ActaHortic.2014.1043.19. doi:10.17660/ActaHortic.2014.1043.19}
}

@misc{ekman2021improved,
  note = {Ekman J. Improved postharvest management of chestnuts. Hort. Innov. Australia Proj. No. CH13005. 2021 Mar;3}
}

@article{de2013design,
  author = {De Kleine, M. E. and Guyer, D. E.},
  title = {Design, Development, and Evaluation of a Single-Stage Combined Chestnut Harvesting and Material Separation Concept},
  journal = {Applied Engineering in Agriculture},
  year = {2013},
  volume = {29},
  number = {6},
  pages = {823-829}
}

@misc{producetech2026silverfox,
  title = {ProduceTech. SilverFox Fruit Harvester},
  url = {https://producetech.com/en/products/silverfox-fruit-harvester/},
  note = {accessed on 1 February 2026}
}

@article{rashid2021a,
  author = {Rashid, M. and Bari, B. S. and Yusup, Y.},
  title = {A comprehensive review of crop yield prediction using machine learning approaches with special emphasis on palm oil yield prediction},
  journal = {IEEE access},
  year = {2021},
  volume = {9},
  pages = {63406-63439}
}

@article{batz2023from,
  author = {Batz, P. and Will, T. and Thiel, S.},
  title = {From identification to forecasting: the potential of image recognition and artificial intelligence for aphid pest monitoring},
  journal = {Frontiers in Plant Science},
  year = {2023},
  volume = {14},
  pages = {1150748}
}

@article{Mamdouh2021,
  author = {Mamdouh, N. and Khattab, A.},
  title = {YOLO-based deep learning framework for olive fruit fly detection and counting},
  journal = {IEEE Access},
  year = {2021},
  volume = {9},
  pages = {84252-84262}
}

@article{alaaudeen2024intelligent,
  author = {Alaaudeen, K. M. and Selvarajan, S. and Manoharan, H.},
  title = {Intelligent robotics harvesting system process for fruits grasping prediction},
  journal = {Scientific Reports},
  year = {2024},
  volume = {14},
  number = {1},
  pages = {2820}
}

@misc{ado2019hru,
note = {Adão, Telmo \& Pádua, Luís \& Pinho, Tatiana M. \& Hruška, Jonáš \& Sousa, A. \& Sousa, Joaquim \& Morais, R. \& Peres, Emanuel. (2019). Multi-Purpose Chestnut Clusters Detection Using Deep Learning: a Preliminary Approach. XLII-3/W8. 1-7. 10.5194/isprs-archives-XLII-3-W8-1-2019. doi:10.5194/isprs-archives-XLII-3-W8-1-2019}
}

@misc{sun20232023,
  note = {Sun, Yifei, Zhenbang Hao, Zhanbao Guo, Zhenhu Liu, and Jiaxing Huang. 2023. "Detection and Mapping of Chestnut Using Deep Learning from High-Resolution UAV-Based RGB Imagery" Remote Sensing 15, no. 20: 4923}
}

@misc{arakawa2024tanaka,
  note = {Arakawa, T., Tanaka, T. S. T., \& Kamio, S. (2024). Detection of on-tree chestnut fruits using deep learning and RGB unmanned aerial vehicle imagery for estimation of yield and fruit load. Agronomy Journal, 116, 973–981}
}

@misc{mccool20162016,
note = {Mccool, Chris \& Sa, Inkyu \& Dayoub, Feras \& Lehnert, Christopher \& Upcroft, Ben \& Perez, Tristan. (2016). Visual Detection of Occluded Crop: for automated harvesting. 10.1109/ICRA.2016.7487405. doi:10.1109/ICRA.2016.7487405}
}

@misc{mu2058a,
  author = {Mu, X. and Lu, Y. and Deng, B.},
  title = {A comparative benchmark of real-time detectors for blueberry detection towards precision orchard management},
  year = {2025},
  howpublished = {arXiv:2509.20580}
}

@article{liao2025yolo,
  author = {Liao, Y. and Li, L. and Xiao, H.},
  title = {YOLO-MECD: citrus detection algorithm based on YOLOv11},
  journal = {Agronomy},
  year = {2025},
  volume = {15},
  number = {3},
  pages = {687}
}

@article{allmendinger2025assessing,
  author = {Allmendinger, A. and Saltık, A. O. and Peteinatos, G. G.},
  title = {Assessing the capability of YOLO-and transformer-based object detectors for real-time weed detection},
  journal = {Precision Agriculture},
  year = {2025},
  volume = {26},
  number = {3},
  pages = {52}
}

@misc{dutta2019the,
  note = {Dutta A, Zisserman A. The VIA annotation software for images, audio and video[C]//Proceedings of the 27th ACM international conference on multimedia. 2019: 2276-2279}
}

@inproceedings{zhang2021a,
  author = {Zhang, Y. and Li, X. and Wang, F.},
  title = {A comprehensive review of one-stage networks for object detection},
  booktitle = {2021 IEEE International Conference on Signal Processing, Communications and Computing (ICSPCC)},
  year = {2021},
  pages = {1-6},
  note = {IEEE}
}

@misc{redmon2016you,
  note = {Redmon J, Divvala S, Girshick R, et al. You only look once: Unified, real-time object detection[C]//Proceedings of the IEEE conference on computer vision and pattern recognition. 2016: 779-788}
}

@misc{glenn2024ultralytics,
  title = {Glenn Jocher, Jing Qiu. Ultralytics YOLO11 [EB/OL]. (2024) [2026-02-05]},
  url = {https://github.com/ultralytics/ultralytics}
}

@misc{tian2025yolov12,
  note = {Tian Y, Ye Q, Doermann D. Yolov12: Attention-centric real-time object detectors[J]. arXiv preprint arXiv:2502.12524, 2025}
}

@misc{lei2025yolov13,
  note = {Lei M, Li S, Wu Y, et al. YOLOv13: Real-Time Object Detection with Hypergraph-Enhanced Adaptive Visual Perception[J]. arXiv preprint arXiv:2506.17733, 2025}
}

@misc{zhu2010deformable,
  note = {Zhu X, Su W, Lu L, Li B, Wang X, and Dai J. "Deformable DETR: Deformable transformers for end-to-end object detection". arXiv preprint arXiv:2010.04159 (2021)}
}

@misc{zhao2024detrs,
  note = {Zhao Y, Lv W, Xu S, et al. Detrs beat yolos on real-time object detection[C]//Proceedings of the IEEE/CVF conference on computer vision and pattern recognition. 2024: 16965-16974}
}

@misc{lv2024rt,
  note = {Lv, Wenyu, Yian Zhao, Qinyao Chang, Kui Huang, Guanzhong Wang, and Yi Liu. "Rt-detrv2: Improved baseline with bag-of-freebies for real-time detection transformer." arXiv preprint arXiv:2407.17140 (2024)}
}

@misc{wang2024rt,
  note = {Wang, Shuo, Chunlong Xia, Feng Lv, and Yifeng Shi. "RT-DETRv3: Real-time End-to-End Object Detection with Hierarchical Dense Positive Supervision." arXiv preprint arXiv:2409.08475 (2024)}
}

@misc{liao2025rt,
  note = {Liao Z, Zhao Y, Shan X, et al. RT-DETRv4: Painlessly Furthering Real-Time Object Detection with Vision Foundation Models[J]. arXiv preprint arXiv:2510.25257, 2025}
}

@misc{le2025thai,
url = {https://doi.org/10.1007/978-981-96-4285-4\_19},
  note = {Le, NT., Thai, N., Bui, C. (2025). Benchmarking Real-Time Object Detection: Evaluating YOLO and RT-DETR on Speed, Accuracy, and Efficiency. In: Buntine, W., Fjeld, M., Tran, T., Tran, MT., Huynh Thi Thanh, B., Miyoshi, T. (eds) Information and Communication Technology. SOICT 2024. Communications in Computer and Information Science, vol 2351. Springer, Singapore. https://doi.org/10.1007/978-981-96-4285-4\_19. doi:10.1007/978-981-96-4285-4\_19}
}

@article{dang2023yoloweeds,
  author = {Dang, F. and Chen, D. and Lu, Y.},
  title = {YOLOWeeds: A novel benchmark of YOLO object detectors for multi-class weed detection in cotton production systems},
  journal = {Computers and Electronics in Agriculture},
  year = {2023},
  volume = {205},
  pages = {107655}
}

@inproceedings{dang2022deepcottonweeds,
  author = {Dang, F. and Chen, D. and Lu, Y.},
  title = {DeepCottonWeeds (DCW): A novel benchmark of YOLO object detectors for weed detection in cotton production systems},
  booktitle = {2022 ASABE Annual International Meeting},
  year = {2022},
  pages = {1},
  note = {American Society of Agricultural and Biological Engineers}
}

@misc{sapkota2040comprehensive,
  note = {Sapkota R, Meng Z, Churuvija M, et al. Comprehensive performance evaluation of yolov12, yolo11, yolov10, yolov9 and yolov8 on detecting and counting fruitlet in complex orchard environments[J]. arXiv preprint arXiv:2407.12040, 2024}
}

@inproceedings{salt2024comparative,
  author = {Saltık, A. O. and Allmendinger, A. and Stein, A.},
  title = {Comparative analysis of yolov9, yolov10 and rt-detr for real-time weed detection},
  booktitle = {European Conference on Computer Vision},
  year = {2024},
  pages = {177-193},
  note = {Cham: Springer Nature Switzerland}
}

@misc{deng2025development,
  note = {Deng B, Lu Y, Brainard D. Development and Preliminary Evaluation of a Machine Vision‐Guided Smart Sprayer Prototype Toward Precision Vegetable Weeding[J]. Journal of Field Robotics, 2025}
}

@article{zhou20243d,
  author = {Zhou, L. and Jin, S. and Wang, J.},
  title = {3D positioning of Camellia oleifera fruit-grabbing points for robotic harvesting},
  journal = {Biosystems Engineering},
  year = {2024},
  volume = {246},
  pages = {110-121}
}

@article{ge2023three,
  author = {Ge, Y. and Xiong, Y. and From, P. J.},
  title = {Three-dimensional location methods for the vision system of strawberry-harvesting robots: development and comparison},
  journal = {Precision Agriculture},
  year = {2023},
  volume = {24},
  number = {2},
  pages = {764-782}
}

@article{chen2024key,
  author = {Chen, Z. and Lei, X. and Yuan, Q.},
  title = {Key technologies for autonomous fruit-and vegetable-picking robots: A review},
  journal = {Agronomy},
  year = {2024},
  volume = {14},
  number = {10},
  pages = {2233}
}
\end{document}